%% file: main.tex
\documentclass[10pt,twocolumn,letterpaper]{article}

\usepackage{arxiv}              % To produce the CAMERA-READY version


\definecolor{arxivblue}{rgb}{0.21,0.49,0.74}
\definecolor{linkcolor}{RGB}{255,0,0}
\definecolor{urlcolor}{RGB}{255,105,180}
\usepackage[pagebackref,breaklinks,colorlinks=true,allcolors=arxivblue,linkcolor=linkcolor,urlcolor=urlcolor]{hyperref}

\usepackage{xspace}
\usepackage{amssymb}
\usepackage{amsmath}
\usepackage{amsfonts}
\usepackage{wrapfig}
\usepackage{algorithm}
\usepackage{algorithmic}

\usepackage{multicol}
\usepackage{multirow}
\usepackage{makecell}
\usepackage{wrapfig}
\usepackage{tabularx}
\usepackage{adjustbox}

\usepackage{pifont}
\usepackage{color}
\usepackage{colortbl}

\definecolor{lightgray}{rgb}{0.8, 0.8, 0.8}
\definecolor{lgray}{rgb}{0.66, 0.66, 0.66}

\definecolor{lblu_tab}{RGB}{225, 235, 246}
\definecolor{orange_vitad}{RGB}{222, 131, 68}
\definecolor{blue_vitad}{RGB}{106, 153, 208}
\definecolor{trajectory_green}{RGB}{126, 171, 85}
\definecolor{trajectory_yellow}{RGB}{245, 194, 66}
\definecolor{tab_others}{RGB}{235, 235, 235}
\definecolor{tab_ours}{RGB}{225, 235, 246}

\definecolor{whit_tab}{RGB}{255, 255, 255}
\definecolor{gray_tab}{RGB}{246, 246, 246}
\definecolor{oran_tab}{RGB}{252, 242, 237}
\definecolor{blue_tab}{RGB}{227, 240, 251}

\def\method{FlexControl}

\title{\method: Computation-Aware ControlNet with Differentiable Router for Text-to-Image Generation}

\author{
  \vspace{4pt}
  Zheng Fang$^1$
  ~~ Lichuan Xiang$^{1,2}$ 
  ~~ Xu Cai$^2$ 
  ~~ Kaicheng Zhou$^2$
  ~~ Hongkai Wen$^1$ %\\
   \\ 
   \normalsize 
   $^1$University of Warwick %Department of Computer Science, University of Warwick, Coventry, UK 
   ~~ $^2$Collov Labs \\
   \normalsize
   zheng.fang.6, lichuan.xiang.3, hongkai.wen@warwick.ac.uk, \\ 
   \normalsize
   caitree@gmail.com, caseyz@collov.com
}
\begin{document}

\twocolumn[{
\renewcommand\twocolumn[1][]{#1}
\maketitle
\begin{center}
    \vspace{-10pt}
    \includegraphics[width=1.0\linewidth]{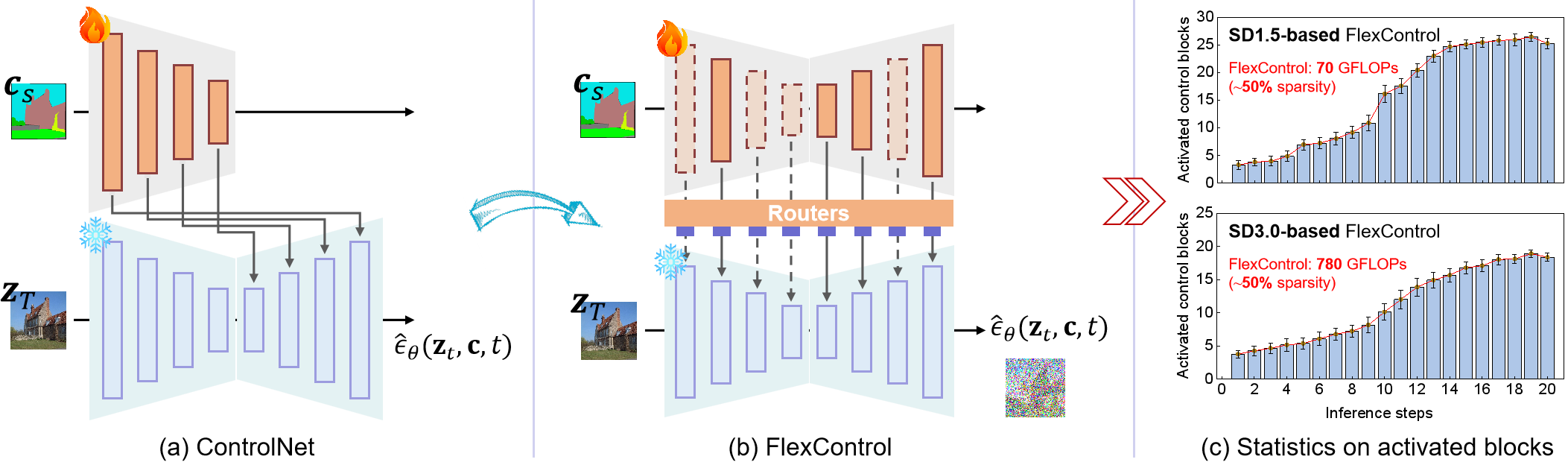}
    \vspace{-20pt}
    \captionsetup{type=figure}
    \caption{%
        \textbf{Dynamically inject conditional controls for image generation based on timestep and specific sample.} (a) The architecture of the ControlNet. (b) The architecture of the proposed FlexControl. (c) Statistics on the number of activated control blocks of the FlexControl at each denoising step. Here, ``50\% spasity" indicates that the number of floating-point operations (FLOPs) of activated blocks is limited to 50\% of the trainable branch.
    }
    \label{fig1}
    % \vspace{-1pt}
\end{center}
}]
\maketitle

\input{sec/0_abstract}    
\input{sec/1_intro}
\input{sec/2_related_work}
\input{sec/3_method}
\input{sec/4_experiment}

\input{sec/5_conclusion}

{
    \small
    \bibliographystyle{ieeenat_fullname}
    \bibliography{main}
}

\input{sec/6_appendix}

\end{document}

%% file: sec/0_abstract.tex
\begin{abstract}
ControlNet offers a powerful way to guide diffusion‐based generative models, yet most implementations rely on ad-hoc heuristics to choose which network blocks to control—an approach that varies unpredictably with different tasks. To address this gap, we propose FlexControl, a novel framework that copies all diffusion blocks during training and employs a trainable gating mechanism to dynamically select which blocks to activate at each denoising step. With introducing a computation-aware loss, we can encourage control blocks only to activate when it benefit the generation quality. By eliminating manual block selection, FlexControl enhances adaptability across diverse tasks and streamlines the design pipeline, with computation-aware training loss in an end-to-end training manner. Through comprehensive experiments on both UNet (\textit{e.g.},~SD1.5) and DiT (\textit{e.g.},~SD3.0), we show that our method outperforms existing ControlNet variants in certain key aspects of interest. As evidenced by both quantitative and qualitative evaluations, FlexControl preserves or enhances image fidelity while also reducing computational overhead by selectively activating the most relevant blocks. These results underscore the potential of a flexible, data‐driven approach for controlled diffusion and open new avenues for efficient generative model design. The code will soon be available at \url{https://github.com/Anonymousuuser/FlexControl}.
\end{abstract}

%% file: sec/1_intro.tex
\section{Introduction}
\label{sec:intro}

Diffusion-based image generation models have recently gained widespread acceptance in the art and design community, not only for their high-quality, photo-realistic image generation but also due to the transformative capabilities of from controllable unit like ControlNet \cite{zhang2023adding}, T2I-Adapter \cite{mou2024t2i}, \textit{etc}, which enables users to create images under diverse conditions(\textit{e.g.}, layout, pose, shape, and form), allows generated image that satisfied various real-world demand.

Despite its growing popularity, ControlNet methods typically rely on multiple design hyperparameters—such as choosing which network block to control for improved fidelity and adherence to input conditions—without a systematic investigation of their effects. For example, the ControlNet variant based on SD1.5 \cite{SD1.5} replicates encoder blocks and injects control information into the decoder, whereas T2I-Adapter applies control in the encoder. It remains unclear which block‐level configuration is most effective, especially since optimal designs can vary by task. Complicating matters further, ControlNet is often trained on significantly smaller datasets than those used for the diffusion model’s pre‐training, implying that adding too much control could disrupt the pretrained representations and degrade image quality, since insufficient control may fail to deliver the desired guidance. As an evidence, a recent study \cite{ju2024brushnet} highlights that the number and placement of control blocks might significantly affect image quality in tasks such as inpainting. Moreover, in practice, ControlNet pipeline relies on heuristic strategies to decide which timesteps should receive control signals at inference, yet evidence is scarce regarding which approach consistently yields the best results. Collectively, these gaps emphasize the need for a more principled, comprehensive analysis of ControlNet design and inference strategies.

To dynamically adjust control blocks based on timestep and conditional information while maintaining (or even improving) generation quality, we propose FlexControl, a data-driven dynamic control method. Similar to conventional controllable generation methods, as shown in \cref{fig1}(a), we freeze the original diffusion model and copy its parameters to process task-specific conditional images. FlexControl is equipped with a router unit within the control block (see \cref{fig1}(b)) to plan forward routes, activating control blocks only when necessary based on the current latent variable.  In contrast to other controllable generation models, FlexControl customizes the inference path for each input, minimizing potential redundant computations. In summary, our main contributions are as follows:

\paragraph{1. Data-driven dynamic control configuration:
} We introduce an automated router unit that dynamically selects control blocks at each timestep, eliminating the need for exhaustive architecture searches and retraining. Our approach enables: (1) task-adaptive control configurations through end-to-end training, (2) temporally adaptive inference via per-timestep activation decisions, and (3) faster configuration design compared to manual search baselines by removing the need for configuration search and repeated training.
\paragraph{2. Computation-aware controllable generation:} Our approach significantly enhances controllability and image quality while maintaining a similar computational cost to the original ControlNet by introducing a novel computation-aware training loss. Specifically, in the depth-map control task, our method achieves a 6.11 FID improvement and a 6.30\% RMSE reduction. Furthermore, this strategic allocation of computation to control units outperforms brute-force doubling, establishing new Pareto frontiers in the control-quality v.s. compute trade-off.
\paragraph{3. Universal plug-and-play integration:} Our method seamlessly integrates with any dual-stream control-model, introducing minimal additional parameters and zero architectural modifications to host models. It enables flexible switching between full control and efficiency-optimized modes, depending on computational requirements.

%% file: sec/2_related_work.tex
\section{Related Work}
\label{sec:formatting}
%-------------------------------------------------------------------------
\paragraph{Text-to-image diffusion models.}
The diffusion probabilistic model was originally introduced by Sohl-Dickstein \textit{et al.} \cite{sohl2015deep}, which has been successfully applied in the field of image synthesis and achieved impressive results \cite{dhariwal2021diffusion, kingma2021variational, huang2023composer, huang2023reversion, jiang2023avatarcraft, ren2022image}. The Latent Diffusion Models (LDMs) \cite{rombach2022high}, reduce computational demands by transferring the diffusion process from the pixel space to the latent feature space. Such diffusion models \cite{SD1.5, nichol2021glide, podell2023sdxl, rombach2022high, saharia2022photorealistic} typically encode text prompts as potential vectors through pre-trained language models \cite{radford2021learning, raffel2020exploring}, combined with UNet \cite{ronneberger2015u} to predict noise to remove at each timestep. Recent studies explore Transformer-based architectures, which have yielded state-of-the-art results for large-scale text-to-image generation tasks \cite{bao2023all, bao2023one, peebles2023scalable, tu2022maxvit, esser2024scaling}, These frameworks leverage Transformers’ capacity for modeling long-range dependencies and scaling to massive multimodal datasets, enabling breakthroughs in compositional reasoning, dynamic resolution adaptation, and high-fidelity synthesis. However, their reliance on purely textual input—despite advances in cross-modal alignment—still poses challenges for precise spatial or stylistic control.

\paragraph{Controllable diffusion models.}
While state-of-the-art text-to-image models achieve remarkable photorealism, their reliance on inherently low-bandwidth, abstract textual input limits their ability to meet the nuanced and complex demands of real-world artistic and design applications. This underscores the growing need for frameworks like ControlNet \cite{zhang2023adding} and T2I-Adapter \cite{mou2024t2i}, which augment text prompts with spatial or structural constraints (\textit{e.g.}, sketches, depth maps, or poses), enabling finer-grained control over generation to bridge the gap between creative intent and algorithmic output. Recent advancements in controllable text-to-image generation have diversified across methodological approaches. Instance-based methods, such as those by \cite{wang2024instancediffusion, zhou2024migc} enable zero-shot generation of stylized images from a single reference input, prioritizing speed and flexibility. Meanwhile, an improvement in cross-attention constraint, proposed by \cite{chen2024training}, guides generation along desired trajectories by refining latent space interactions. Prompt engineering has also emerged as a lightweight strategy for enhancing controllability, with works like \cite{ju2023humansd, zhang2023controllable, yang2023reco, li2023gligen} optimizing textual or hybrid prompts for fine-grained guidance. Additionally, multi-condition frameworks  \cite{hu2023cocktail, qin2023unicontrol, zhao2024uni, li2025controlnet} integrate auxiliary inputs—such as segmentation maps or depth cues—to complement text prompts, improving alignment with complex user intent. However, while these methods expand generative versatility, many overlook the computational overhead introduced by auxiliary networks, limiting their scalability for real-time applications.

\paragraph{Improving ControlNet efficiency.}
Efforts to enhance ControlNet’s efficacy and efficiency have focused on architectural redesigns, training optimizations, and inference acceleration. ControlNeXt \cite{peng2024controlnext} 
replaces ControlNet’s bulky auxiliary branches with a streamlined architecture and substitutes zero modules with Cross Normalization, slashing learnable parameters by 90\% while maintaining stable convergence. Beyond this, multi-expert diffusion frameworks \cite{lee2024multi, zhang2023improving} tailor denoising operations to specific timesteps, though their computational demands hinder practicality. To reduce inference costs, pruning techniques\cite{fang2023spdm, kim2023architectural, ganjdanesh2024not} trim redundant parameters from pre-trained denoising models, while distillation methods\cite{hsiao2024plug} train lightweight guide models to minimize denoising steps. Inspired by RepVGG  \cite{ding2021repvgg}, RepControlNet \cite{deng2024repcontrolnet} introduces a novel reparameterization strategy: modal-specific adapters modulate features during training, and their weights are later merged with the base network, eliminating auxiliary computations at inference. Unlike prior methods that rely on fixed heuristics, post-hoc pruning, or static architectural modifications, our FlexControl introduces a dynamic, end-to-end trainable framework where block activation is both task-aware and computation-aware. By integrating a gating mechanism with a computational efficiency objective, our approach uniquely balances precision and resource usage, enabling adaptive control across diverse architectures (UNet, DiT) without manual intervention — a paradigm shift from rigid, task-specific designs to flexible, generalizable control.

%% file: sec/3_method.tex
\section{Methodology}

\begin{figure*}[!t]
\centering
	\includegraphics[width=1.0\textwidth]{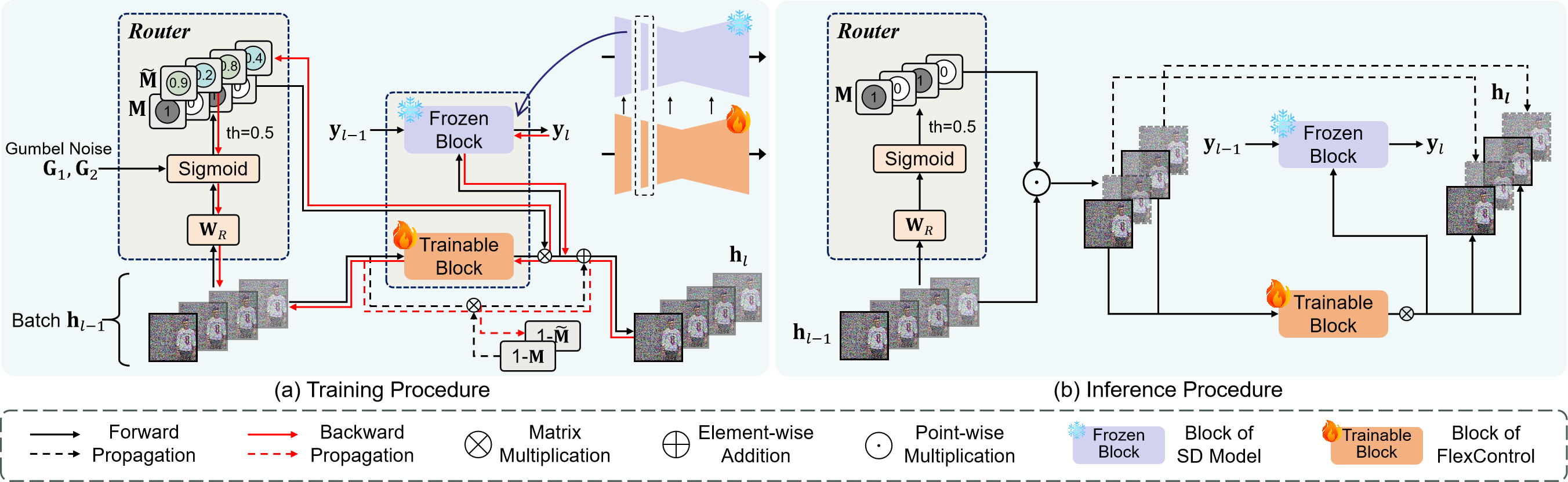}
	\caption{\textbf{Overview of dynamic routing guided by the router unit.} (a) In the training stage, Gumbel noise is added to the discrete mask to assist the gradient backpropagation. (b) In the inference stage, the router unit controls whether to activate the control block and whether to inject conditional control into the frozen block of the backbone according to the input latent variable. Once output the instruction of inactive, the corresponding control block and zero module will be skipped.}
    \label{fig2}
\end{figure*}

\subsection{Preliminaries}

Denoising diffusion probabilistic model (DDPM) \cite{ho2020denoising} aims to approximate the real data distribution $q\left(\mathbf{x}_{0}\right) $ with the learned model distribution $p\left(\mathbf{x}_{0}\right) $ \cite{ho2020denoising}. It contains a forward diffusion process that progressively adds noise to the image and a reverse generation process that synthesizes the image by progressively eliminating noise. Formulaly, the forward process is a $T$-step Markov chain:
\begin{equation}
\begin{split}                   
    &q\left(\mathbf{x}_{1:T}|\mathbf{x}_{0}\right):=\prod_{t=1}^{T}q\left(\mathbf{x}_{t}|\mathbf{x}_{t-1}\right), \\
    &q\left(\mathbf{x}_{t}|\mathbf{x}_{t-1}\right):=\mathcal{N}\left(\mathbf{x}_{t};\sqrt{1-\beta_{t}}\mathbf{x}_{t-1},\beta_{t}\boldsymbol{\mathit{I}}\right),
\end{split}
\end{equation}
where $\left\{\beta_{t}\right\}_{t=0}^{T}$ are the noise schedule, and $\left\{\mathbf{x}_{t}\right\}_{t=0}^{T}$ are latent variables. Let $\alpha_{t}=1-\beta_{t}$, the distribution of $\mathbf{x}_{t}$ for a given $\mathbf{x}_{0}$ can be expressed as:
\begin{equation}
    q\left(\mathbf{x}_{t}|\mathbf{x}_{0}\right):=\mathcal{N}\left(\mathbf{x}_{t};\sqrt{\bar{\alpha}_{t}}\mathbf{x}_{t-1},\left(1-\bar{\alpha}_{t}\right)\boldsymbol{\mathit{I}}\right).
\label{eq:ddpm_forward}
\end{equation}
Here, $\bar{\alpha}_{t}=\prod_{i=0}^{t}\alpha_{i}$ is a differentiable function of timestep $t$, which is determined by the denoising sampler. Therefore, the diffusion training loss can be formulated as:
\begin{equation}
    \mathcal{L}_{\theta}=\mathbb{E}_{\mathbf{x}_{0},t\sim\mathcal{U}\left(t\right),\epsilon\sim\mathcal{N}\left(\mathbf{0},\boldsymbol{\mathit{I}}\right)}\left[w\left(\lambda_t\right)\left\|\hat{\epsilon}_{\theta}\left(\mathbf{x}_t,t\right)-\epsilon\right\|_{2}^{2}\right],
\label{eq:optimization_object}
\end{equation}
where $\epsilon$ denotes a noise vector drawn from a Gaussian distribution, and $\hat{\epsilon}_{\theta}$ refers to the predicted noise at timestep $t$ by denoising model with parameters $\theta$. $w\left(\lambda_t\right)$ is a pre-defined weighted function that takes into the signal-to-noise ratio $\lambda_t$. The reverse process first sample a Gaussian noise $p\left(\mathbf{x}_{T}\right)=\mathcal{N}\left(\mathbf{x}_{T};\mathbf{0},\boldsymbol{\mathit{I}}\right)$, and then proceeding with the transition probability density step by step:
\begin{equation}
\begin{split}
    p_{\theta}\left(\mathbf{x}_{t-1}|\mathbf{x}_{t}\right)&\approx q\left(\mathbf{x}_{t-1}|\mathbf{x}_{t},\mathbf{x}_{0}\right) \\
    &=\mathcal{N}\left(\mathbf{x}_{t-1};\mu_{\theta}\left(\mathbf{x}_{t},\mathbf{x}_{0}\right),\sigma_{t}^{2}\boldsymbol{\mathit{I}}\right),
\end{split}
\end{equation}
where $\mu_{\theta}\left(\mathbf{x}_{t},\mathbf{x}_{0}\right)=\frac{1}{\sqrt{\alpha_t}}\left(\mathbf{x}_{t}-\frac{1-\alpha_{t}}{\sqrt{1-\bar{\alpha}_{t}}}\epsilon_{\theta}\left(\mathbf{x}_{t},t\right)\right)$ and $\sigma_{t}^{2}=\frac{1-\bar{\alpha}_{t-1}}{1-\bar{\alpha}_{t}}\beta_{t}$ are the mean and variance of posterior Gaussian distribution $p_{\theta}\left(\mathbf{x}_{0}\right)$.

In order to improve the efficiency of diffusion model, flow-based optimization strategy \cite{lipman2022flow,liu2022flow,liu2023instaflow} is introduced, which defines the forward process as a straight path between the real data distribution and the standard normal distribution:
\begin{equation}
    \mathbf{x} _t=a_t\mathbf{x} _0+b_t\epsilon. 
\label{eq:fm_forward}
\end{equation}
With \cref{eq:fm_forward}, a vector field $u_t$ is constructed to generate a path $p_t$ between the noise distribution and the data distribution. Meanwhile, the velocity $v$ is parameterized by the weight $\theta$ of a neural network to approximate $u_t$. After variable recombination, the flow matching object can also be formulated as \cref{eq:optimization_object} \cite{SD3}. In the reverse stage, flow matching uses ODE solver for fast sampling:
\begin{equation}
    \mathbf{x}\left(t\right)=\mathbf{x}\left(0\right)+\int_{0}^{t}v_{\theta}\left(\mathbf{x\left(\tau\right),\tau}\right)\mathrm{d}\tau.   
\end{equation}

\subsection{Structure}
\label{sec:structure}

Building on the core design philosophy of ControlNet, we first fix the powerful diffusion model backbone, fine-tune a trainable copy with zero module to learn conditional controls, and then inject the acquired knowledge into the frozen backbone:
\begin{equation}
    \mathbf{y}=\mathcal{F}\left(\mathbf{x};\Theta \right) + \mathcal{Z}\left(\mathcal{F}\left(\mathbf{x}+\mathcal{Z}\left(\mathbf{c};\Theta_{z1}\right);\Theta_{c}\right);\Theta_{z2}\right),
\label{eq:inject_conditional_controls}
\end{equation}
where $\Theta$ and $\Theta_{c}$ are the weight parameters of the original block and the trainable copy respectively, $\mathcal{Z}$ represents zero module and $\mathbf{c}$ is the control element. Instead of just cloning the encoder and adding conditional controls only in the decoder blocks \cite{zhang2023adding}, as shown in \cref{fig1}(a), we copy all blocks of the original diffusion model to generate conditional controls and inject them into the corresponding blocks of the backbone in turn, as shown in \cref{fig1}(b), which is similar to the strategy used in BrushNet \cite{ju2024brushnet}, and we call this structure ControlNet-Large. Although the double branch structure improves the quality of the generated image, it leads to huge redundant computation and multiplies the inference delay. 

To reduce computational redundancy and enhance image generation quality, we propose FlexControl, which introduces a lightweight router unit before each conditional control generation block in the trainable branch.  The router generates a binary mask $\mathcal{M}\in\left\{0,1\right\}^{N}$ from the input latent feature, determining whether the underlining control block needs to be activated.  Specifically, ``0'' indicates inactive, ``1'' indicates activate, and $N$ represents the number of control blocks in the trainable branch. 

The mask generation process of the router is data-driven, enabling independent path planning and adaptive decision-making based on the input latent representation. As shown in \cref{fig2}, during inference, if the router outputs a mask value of ``0'', the conditional mapping skips the next control block until activation is deemed necessary. Taking the $l$-th control block as an example, the computation process can be formulated as:
\begin{equation}
\mathbf{h}_{l}=
\begin{cases}
    \mathcal{F}_{l}\left(\mathbf{h}_{l-1},\mathbf{c},t;\Theta_{c}^{l} \right)
    &\text{if} \ \mathcal{M}_{l}=1 \\
    \mathrm{skip}_{l}\left(\mathbf{h}_{l-1}\right)
    &\text{if} \ \mathcal{M}_{l}=0,
\end{cases}
\label{eq:forward_backward_1}
\end{equation}
where $\mathcal{F}_{l}\left(\cdot\right)$ indicates the $l$-th control block operation with parameter $\Theta_{c}^{l}$, $\mathbf{h}_{l}$ is the output of it at timestep $t$, and $\mathrm{skip}_{l}\left(\cdot\right)$ is used to bypass the current block. Following the design of \cite{zhang2023adding}, we utilize the zero module to transform the latent feature $\mathbf{h}_{l}$ into conditional control:
\begin{equation}
\mathbf{y}_{c}^{l}=
\begin{cases}
    \mathcal{Z}_{l}\left(\mathbf{h}_{l};\Theta_{z}^{l}\right)
    &\text{if} \ \mathcal{M}_{l}=1 \\
    \mathrm{N/A}
    &\text{if} \ \mathcal{M}_{l}=0.
\end{cases}
\label{eq:forward_backward_2}
\end{equation}
Here, $\mathbf{y}_{c}^{l}$ denotes the conditional control incorporated into the feature space of the backbone.

{\bf Remark:} The above designed router is in fact lightweight, accounting for less than 1\% of the parameters of the overall model. Since the skipped parameters are excluded from tensor computation during inference, FlexControl barely introduces computational burden by adaptively adjusting the number of active control blocks. See the detailed inference process in \cref{alg:inference} in \cref{app:pseudo}.

\subsection{Router unit design}

As illustrated earlier, the router unit is lightweight and plug-and-play to any diffusion architecture. However, given the differences between UNet and DiT, we will discuss the implementation of the router on these two commonly used architectures separately.

\paragraph{Router for UNet-based architecture.}
The output of UNet block is a multi-channel spatial feature. Given input $\mathbf{h}\in\mathbb{R}^{C\times H\times W}$, the router unit first transforms the spatial feature into linear feature $\mathbf{h}^{'}\in\mathbb{R}^{C}$ through the downsampling layer, we use global average pooling (GAP) in implementation, and then the MLP layer with weight $\mathbf{W}\in \mathbb{R}^{C\times 1}$ maps the linear feature into a scalar $\mathcal{K}$:
\begin{equation}
   \mathcal{K}=\mathrm{MLP}\left(\mathrm{GAP}\left(\mathbf{h}\right)\right).  
\end{equation}
Henceforth, we compute a new scalar $\mathcal{K}^{'}$ by restricting the value of $\mathcal{K}$ to the interval $\left(0,1\right)$ through the Sigmoid function. In order to convert $\mathcal{K}^{'}$ into a binary coding, we introduce a threshold discriminator to control the generation of the mask $\mathcal{M}$ by a preset threshold $\mathcal{T}$ (0.5 by default):
\begin{equation}
\mathcal{M}=
\begin{cases}
    1 & \text{if} \ \mathcal{K}^{'}> \mathcal{T} \\
    0 & \text{if} \ \mathcal{K}^{'}\le \mathcal{T}.
\end{cases}
\label{eq:discriminator}
\end{equation}
We multiply the mask $\mathcal{M}$ and the output latent feature to zero out the corresponding control block and zero module. It can be seen from the above description that the mask $\mathcal{M}$ is learned from the latent variable $\mathbf{h}$. Since timestep embedding is introduced into the blocks of the diffusion model during the generation of $\mathbf{h}$, the output of the router is also affected by the sampled timesteps.

\paragraph{Router for DiT-based architecture.}

For the router applied in DiT, we conduct feature analysis from multiple perspectives. Specifically, we perform both global and local feature encoding on the latent variable $\mathbf{h}\in\mathbb{R}^{N\times C}$ output by the Transformer block \cite{rao2021dynamicvit, rao2023dynamic}. The detailed encoding process is as follows:
\begin{equation}
    \mathbf{h}^{global}=\mathrm{MLP}^{global}\left(\mathrm{AVG}_{dim=1}\left(\mathbf{h}\right)\right),
\label{eq:mlp_global}
\end{equation}
\begin{equation}
    \mathbf{h}^{local}=\mathrm{MLP}^{local}\left(\mathrm{AVG}_{dim=2}\left(\mathbf{h}\right)\right).
\label{eq:mlp_local}
\end{equation}
From \cref{eq:mlp_global,eq:mlp_local}, the encoding process for global and local features primarily consists of two steps. First, feature fusion is performed across all tokens and hidden channels using the function $\mathrm{AVG\left(\cdot\right)}$, which is implemented via average pooling along different dimensions of latent variable. This yields the global feature $\mathbf{z}^{global}\in \mathbb{R}^{C}$ and local feature $\mathbf{z}^{local}\in \mathbb{R}^{N}$. Second, the embedding dimensions of $\mathbf{z}^{global}$ and $\mathbf{z}^{local}$ are aligned through an MLP layer and reduced to $\mathcal{O}$, which is set to $C/64$ by default. Intuitively, the local feature captures token-specific information, while the global feature encodes potential relationships between tokens. We then merge these global and local features to form a new feature representation:
\begin{equation}
    \mathbf{h}^{mix}=\alpha_{1}\cdot \mathbf{h}^{global}+\alpha_{2}\cdot \mathbf{h}^{local}.
\end{equation}
In the above equation, $\alpha_{1}$ and $\alpha_{2}$ are weight factors that balance the influence of global and local features, both set to 0.5 by default. The fused feature variable $\mathbf{h}_{l-1}^{mix}\in \mathbb{R}^{\mathcal{O}}$ is then passed through an MLP layer to produce $\mathcal{K}$. At the end, $\mathcal{K}$ is processed through a Sigmoid layer followed by the threshold discriminator described in \cref{eq:discriminator}, resulting in the router mask $\mathcal{M}$.

\subsection{End-to-end training}

\paragraph{Differentiable learning of router.}

To enable end-to-end training via gradient descent, we address the discrete, non-differentiable nature of the mask by incorporating Gumbel noise into the Sigmoid activation function. This allows the discrete mask $\mathcal{M}$ to be approximated by the differentiable Gumbel-Sigmoid version $\widetilde{\mathcal{M}}$ during training:
\begin{equation}    
    \widetilde{\mathcal{M}}_{l}=\mathrm{Sigmoid}\bigg(\frac{\mathcal{R}_{l}\left(\mathbf{h}_{l-1};\Theta_{\mathcal{R}}^{l}\right)+G_{1}-G_{2}}{\mathcal{TP}}\bigg),
\end{equation}
where $G_{1}$, $G_{2}$ $\sim$ Gumbel$(0, 1)$, $\mathcal{TP}$ denotes the temperature hyperparameter (5 by default), $\mathcal{R}\left(\cdot\right)$ denotes tensor computations in the router unit parametered by $\Theta_{\mathcal{R}}$.

To this end, we employ different mask schemes during the forward and backward passes:
\begin{equation}
\mathbf{h}_{l}=
\begin{cases}
    \mathcal{F}_{l}\cdot\mathcal{M}_{l}+\mathrm{skip}_{l}\left(\mathbf{h}_{l-1}\right)\cdot\left(1-\mathcal{M}_{l}\right)
    &\text{if Forward} \\
    \mathcal{F}_{l}\cdot\widetilde{\mathcal{M}}_{l}+\mathrm{skip}_{l}\left(\mathbf{h}_{l-1}\right)\cdot\big(1-\widetilde{\mathcal{M}}_{l}\big)
    &\text{if Backward}.
\end{cases}
\label{eq:redefined_forward_backward_1}
\end{equation}
Meanwhile, the computation process of the zero module is adjusted accordingly:
\begin{equation}
\mathbf{y}_{c}^{l}=
\begin{cases}
    \mathcal{Z}_{l}\left(\mathbf{h}_{l};\Theta_{z}^{l}\right)\cdot \mathcal{M}_{l}
    &\text{if Forward} \\
    \mathcal{Z}_{l}\left(\mathbf{h}_{l};\Theta_{z}^{l}\right)\cdot \widetilde{\mathcal{M}}_{l}
    &\text{if Backward}.
\end{cases}
\label{eq:redefined_forward_backward_2}
\end{equation}

{\bf Remark:} As can be seen in \cref{eq:redefined_forward_backward_1,eq:redefined_forward_backward_2} during training, the blockwise routing differs from the inference process displayed in \cref{eq:forward_backward_1,eq:forward_backward_2}: during training, we retain all blocks to ensure proper back-propagation, rather than skipping blocks as done during inference. %slightly 

\paragraph{Computation-aware training loss.}

\begin{figure*}[!t]
    \centering
	\includegraphics[width=1.0\textwidth]{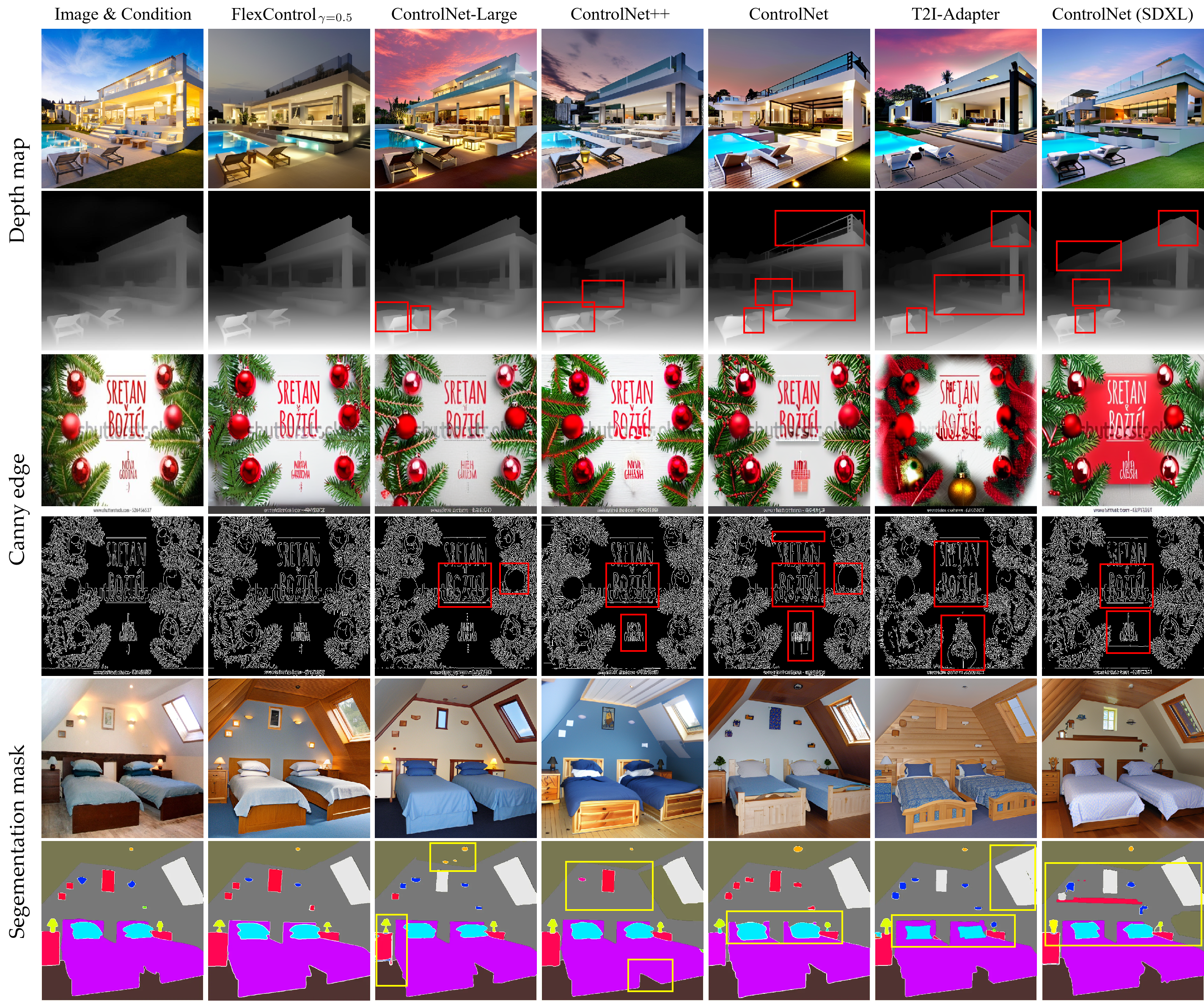}
    \caption{
    \textbf{Qualitative comparison of controllable generation methods.} FlexControl achieves higher fidelity and structure preservation across Depth Map, Canny Edge, and Segmentation Mask conditions, reducing distortions (boxes) seen in other methods. It better aligns with input conditions while maintaining visual quality.
    }
    \label{diode}
\end{figure*}

\begin{table*}[!t]
    \centering
    \resizebox{\linewidth}{!}{
    \begin{tabular}{c|c|cc|cc|cc|cc} 
        \toprule[1.0px]
        \textbf{Method}&\textbf{Base Model}& \multicolumn{2}{c|}{\textbf{Depth Map}}&\multicolumn{2}{c|}{\textbf{Canny Edge}}& \multicolumn{2}{c|}{\textbf{Seg. Mask}}& \multicolumn{2}{c}{\textbf{\#Average}}\\ & & \textbf{FID} $\downarrow$& \textbf{CLIP\_score} $\uparrow$& \textbf{FID} $\downarrow$& \textbf{CLIP\_score} $\uparrow$& \textbf{FID} $\downarrow$&\textbf{CLIP\_score} $\uparrow$& \textbf{FID} $\downarrow$&\textbf{CLIP\_score} $\uparrow$\\
        \midrule
        ControlNet \cite{zhang2023adding} &SDXL& 19.90& 0.3224& 22.07& 0.2657& 26.95& 0.2495& 22.97& 0.2792\\
        T2I-Adapter \cite{mou2024t2i} &SDXL& 19.74& 0.3197& 22.91& 0.2614& 27.54& 0.2501& 23.40& 0.2771\\
        \midrule
        GLIGEN \cite{li2023gligen} &SD1.4& 18.36& 0.3175& 19.01& 0.2520& 23.79& 0.2490& 20.39& 0.2728\\
        T2I-Adapter \cite{mou2024t2i} &SD1.5& 22.52& 0.3146& 16.74& 0.2598& 24.65& 0.2494& 21.30& 0.2728\\
        ControlNet \cite{zhang2023adding} & SD1.5& 17.76& 0.3245& 15.23& 0.2613& 21.33& 0.2531& 18.11& 0.2796\\ 
        ControlNet++ \cite{li2025controlnet} &SD1.5& 16.66& 0.3209& 17.23& 0.2598& 19.89& 0.2640& 17.93& 0.2816\\
        ControlNet-Large &SD1.5& \textcolor{blue}{12.45}& \textcolor{blue}{0.3492}&
        \textcolor{blue}{12.92}& \textcolor{red}{0.2789}& \textcolor{blue}{16.78}& \textcolor{blue}{0.2796}& \textcolor{blue}{14.05}& \textcolor{blue}{0.3026}\\
        \midrule
        \rowcolor{gray!20}
        \textbf{FlexControl}$_{\gamma=0.5} $&SD1.5& \textcolor{red}{11.65}& \textcolor{red}{0.3498}& \textcolor{red}{11.37}& \textcolor{blue}{0.2778}& \textcolor{red}{14.80}& \textcolor{red}{0.2842}& \textcolor{red}{12.61}& \textcolor{red}{0.3039}\\ 
        \bottomrule[1.0px]
    \end{tabular}
    }
    \caption{ 
    \textbf{Quantitative comparison of FlexControl with state-of-the-art methods.} We report FID ($\downarrow$) and CLIP score ($\uparrow$) on different conditioning types: Depth Map, Canny Edge, and Segmentation Mask. Lower FID indicates better image quality, while higher CLIP score reflects better alignment with textual prompts. The best results are highlighted in \textcolor{red}{red}, while the second-best results are shown in \textcolor{blue}{blue}. FlexControl achieves the best overall performance, demonstrating superior fidelity and semantic alignment.
    }
    \label{tab:table1}
\end{table*}

Following standard controllable generation methods, our training dataset $\mathcal{D}$ contains triples of the original image $x$, spatial conditioning control $\mathbf{c}_{s}$, and text prompt $\mathbf{c}_{t}$. The diffusion loss of FlexControl is formulated as:
\begin{equation}
    \mathcal{L}_{\mathbf{SD}}=\mathbb{E}_{\mathbf{x}_{0},\mathbf{c}_{t},\mathbf{c}_{s},t,\epsilon\sim \mathcal{N}\left(\mathbf{0},\boldsymbol{\mathit{I}}\right)}\left[\left\|\hat{\epsilon}_{\theta}\left(\mathbf{x}_{t},\mathbf{c}_{t},\mathbf{c}_{s},t\right)-\epsilon \right\|_{2}^{2}\right].
\label{eq:losssd}
\end{equation}
FlexControl aims to activate the optimal control blocks and inject conditional mapping into the backbone network for efficient image generation. In addition to the regular diffusion loss $\mathcal{L}_{\mathbf{SD}}$, we introduce a cost loss $\mathcal{L}_{\mathbf{C}}$ to regulate resource consumption to the desired sparsity $\gamma$, which measures the proportion of floating-point operations (FLOPs):
\begin{equation} 
\mathcal{L}_{\mathbf{C}}=\frac{1}{\left|\mathcal{D}_{\mathrm{bs}}\right|}\sum_{d\in\mathcal{D}_{\mathrm{bs}}}\left(\frac{F^{\mathrm{Flex}}_{t_d}\left(d\right)}{F^{\mathrm{Large}}_{t_d}\left(d\right)}-\gamma\right)^{2},
\label{eq:losscost}
\end{equation}
where $\mathcal{D}_{\mathrm{bs}}$ represents the current batch samples, $t_d\in \left[0,T\right]$ is the uniformly sampled timestep for sample $d$. $F_{t}\left(d\right)$ denotes FLOPs of the trainable branch at sampled timestep, and superscripts $\mathrm{Flex}$ and $\mathrm{Large}$ respectively denote FlexControl and ControlNet-Large. We combine $\mathcal{L}_{\mathbf{SD}}$ and $\mathcal{L}_{\mathbf{C}}$ to bring out the final optimization goal,
\begin{equation} 
 \mathcal{L}_{\theta}= \mathcal{L}_{\mathbf{SD}}+\lambda_{\mathbf{C}}\cdot\mathcal{L}_{\mathbf{C}},
 \label{eq:losstheta}
\end{equation}
where $\lambda_{\mathbf{C}}$ is the hyperparameter that controls the influence of loss $\mathcal{L}_{\mathbf{C}}$. See the detailed training process in \cref{alg:training} in \cref{app:pseudo}.

%% file: sec/4_experiment.tex
\section{Experiment}

We evaluate FlexControl against state-of-the-art methods across different image conditions: depth maps (MultiGen-20M, \cite{zhao2024uni}), canny edges (LLAVA-558K, \cite{liu2024visual}), segmentation masks (ADE20K, \cite{zhou2017scene}), and \textit{etc}.

\subsection{Quantitative comparison}

\paragraph{Comparison of image quality.}

To evaluate the impact of dynamic controllable generation on image quality, we compare the FID metrics of different methods across multiple conditional generation tasks (\cref{tab:table1}). We set $\gamma$ to 0.5, aligning FlexControl’s FLOPs with ControlNet’s. Our model achieves superior FID results across all conditions, outperforming existing methods. We also examine ControlNet-Large, which replicates the SD model as an additional network. Although its larger parameter count enhances conditional feature extraction and control, its performance remains inferior to FlexControl$_{\gamma=0.5}$. This confirms that adaptive control—selectively applying conditions instead of enforcing them across all blocks and timesteps—maximizes controllability. Beyond spatial conditions, we assess text influence using CLIP score. As shown in \cref{tab:table1}, FlexControl$_{\gamma=0.5}$ outperforms other methods, demonstrating that precise control enhances spatially guided generation without compromising text-guided synthesis. Additionally, we evaluate ControlNet and T2I-Adapter on the SDXL backbone \cite{podell2023sdxl}, revealing that a larger backbone does not necessarily improve image quality.

\begin{table}
    \centering
    \resizebox{\linewidth}{!}{
    \begin{tabular}{c|c|c|c|c} 
        \toprule[1.0px]
        \textbf{Method}  &\textbf{Base Model}& \textbf{Depth Map}& \textbf{Canny Edge}&\textbf{Seg. Mask}\\ & & \textbf{(RMSE $\downarrow$)}& \textbf{(SSIM $\uparrow$)}&\textbf{(mIoU $\uparrow$)}\\ 
        \midrule
        ControlNet &SDXL& 0.4001& 0.4178& 0.2058\\
        T2I-Adapter &SDXL& 0.3976& 0.3969& 0.1912\\ 
        \midrule
        GLIGEN &SD1.4& 0.3882& 0.4226& 0.2076\\
        T2I-Adapter &SD1.5& 0.4840& 0.4622& 0.1839\\
        ControlNet & SD1.5& 0.2988& 0.5197& 0.2764\\
        ControlNet++ &SD1.5& 0.2832& 0.5436& 0.3435\\
        ControlNet-Large &SD1.5& \textcolor{blue}{0.2372}& \textcolor{red}{0.5642}& \textcolor{blue}{0.3668}\\ 
        \midrule
        \rowcolor{gray!20}
        \textbf{FlexControl}$_{\gamma=0.5}$&SD1.5& \textcolor{red}{0.2358}& \textcolor{blue}{0.5612}& \textcolor{red}{0.3751}\\ 
        \bottomrule[1.0px]
    \end{tabular}
    }
    \caption{
        \textbf{Controllability comparison across different conditioning types.} 
        We report RMSE (↓) for Depth Map and SSIM (↑) for Canny Edge and mIoU (↑) for Seg. Mask.
        The best and second-best results are highlighted in \textcolor{red}{red} and \textcolor{blue}{blue}. FlexControl achieve similar but slightly better controllability than ControlNet-Large with only \textbf{half} activation blocks.
    }
    \label{tab:table2}
\end{table}

\paragraph{Comparison of controllability.}
We exam generation controllability in detail by comparing results across different spatial conditions. ControlNet and its variants generally achieve stronger controllability than other existing methods. Within a similar computational budget, FlexControl further improves controllability across various conditions.  Numerically, FlexControl reduces RMSE by 6.30\% and 4.74\% compared to ControlNet and ControlNet++ on the depth map task. For canny edge and segmentation mask, FlexControl shows improvements of 4.15\%/1.76\% in SSIM and 9.87\%/3.16\% in mIoU, respectively. Moreover, our method outperforms ControlNet-Large on both the depth map and segmentation mask datasets, and achieves similar performance on the canny edge task. Similarly, we show the results of the SDXL-based ControlNet and T2I-Adapter show only marginal improvements for specific tasks.

\begin{table}
    \centering
    \resizebox{\linewidth}{!}{
    \begin{tabular}{c|c|c|c|c} 
        \toprule[1.0px]
        \textbf{Method}  &\textbf{Base Model}& \textbf{Param.}& \textbf{FLOPs}&\textbf{Speed}\\ 
        \midrule
        ControlNet &SD1.5&\textcolor{red}{0.36 G} &\textcolor{blue}{233 G} &\textcolor{blue}{{5.23\scriptsize{$\pm$0.07}} it/s}\\
        ControlNet-Large &SD1.5&0.72 G &561 G & 4.02{\scriptsize{$\pm$0.05}} it/s\\
        \rowcolor{gray!10}
        \textbf{FlexControl}$_{\gamma=0.7}$&SD1.5&0.73 G &393 G &4.94{\scriptsize{$\pm$0.07}} it/s\\ 
        \rowcolor{gray!10}
        \textbf{FlexControl}$_{\gamma=0.5} $&SD1.5&0.73 G &280 G &\textcolor{blue}{{5.21\scriptsize{$\pm$0.12}} it/s}\\
        \rowcolor{gray!10}
        \textbf{FlexControl}$_{\gamma=0.3} $&SD1.5&0.73 G &\textcolor{red}{168 G} &\textcolor{red}{{5.64\scriptsize{$\pm$0.12}} it/s}\\ 
        \midrule
        ControlNet &SD3.0& \textcolor{red}{1.06 G}&3.25 T &48.34{\scriptsize{$\pm$1.78}} s/it\\
        ControlNet-Large &SD3.0& 2.02 G&6.22 T &59.46{\scriptsize{$\pm$1.82}} s/it\\
        \rowcolor{gray!10}
        \textbf{FlexControl}$_{\gamma=0.7}$&SD3.0& 2.03 G&4.35 T &52.15{\scriptsize{$\pm$2.86}} s/it\\ 
        \rowcolor{gray!10}
        \textbf{FlexControl}$_{\gamma=0.5}$&SD3.0& 2.03 G&\textcolor{blue}{3.11 T} &\textcolor{blue}{45.74{\scriptsize{$\pm$3.25}} s/it}\\
        \rowcolor{gray!10}
        \textbf{FlexControl}$_{\gamma=0.3}$&SD3.0& 2.03 G&\textcolor{red}{1.86 T} &\textcolor{red}{40.84{\scriptsize{$\pm$3.09}} s/it}\\
        \bottomrule[1.0px]
    \end{tabular}
    }
    \caption{
        \textbf{Complexity comparison on SD1.5 and SD3.0.} 
        We compare model parameters, FLOPs, and inference speed (it/s (↑), iterations per second and s/it (↓), seconds per iteration). The best values are highlighted in \textcolor{red}{red}, while the second-best values are shown in \textcolor{blue}{blue}. FlexControl significant reduce overall FLOPs and inference time from ControlNet-Large.
    }
    \label{tab:table3}
\end{table}

\subsection{Qualitative comparison}

In \cref{diode}, we compare different methods across depth map, canny edge, and segmentation mask tasks. All models use SD1.5 as the backbone, except for the last two columns. FlexControl consistently outperforms others in visual quality and spatial/text alignment.For depth maps, FlexControl produces smoother transitions and more natural textures. Under canny edge conditions, it better preserves edge fidelity and fine details. For segmentation masks, it enhances mask reconstruction and visual consistency. These results demonstrate FlexControl’s ability to selectively inject control information into relevant diffusion backbone blocks based on timestep and input characteristics, improving image fidelity.Finally, we compare against ControlNet and ControlNet-Large. While ControlNet-Large benefits from a larger control network for improved generation and condition alignment, FlexControl surpasses it in both accuracy and visual fidelity, showcasing the strength of our approach.

\begin{figure}[!t]
    \centering
        \includegraphics[width=1.0\columnwidth]{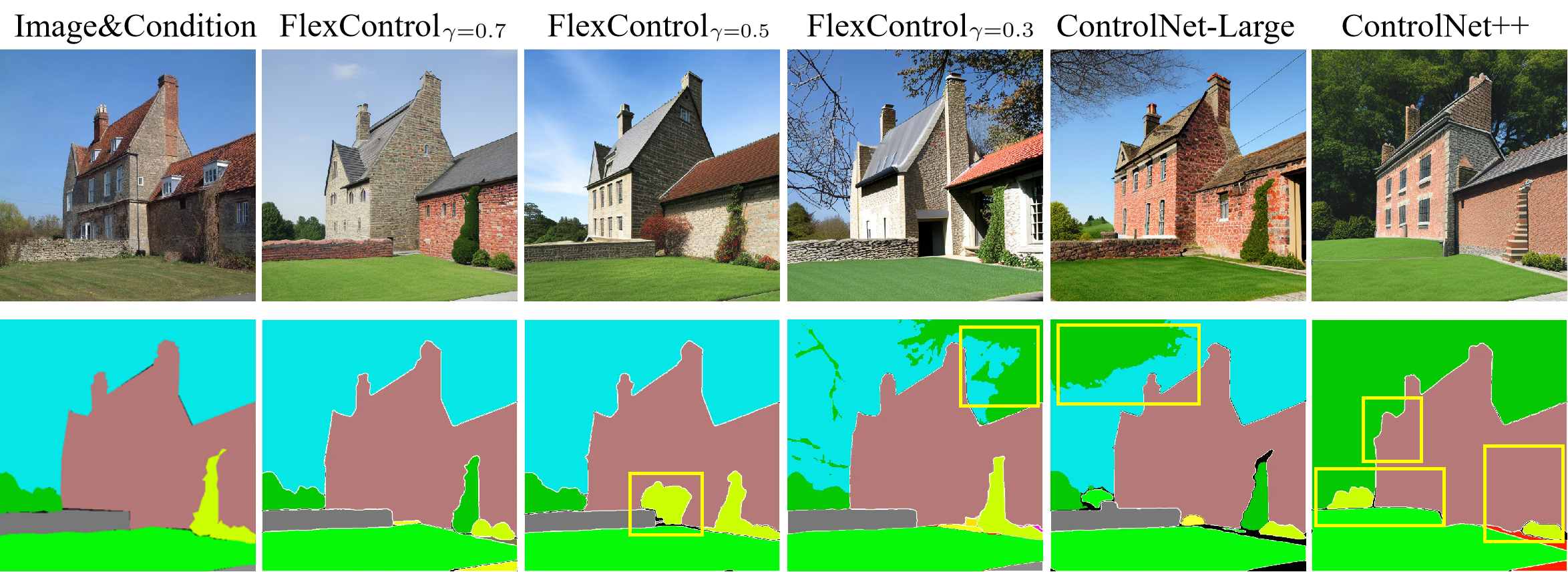}
        \caption{
        \textbf{Comparison of FlexControl and existing methods on SD1.5 for semantic consistency.} 
        FlexControl achieves better semantic alignment and structure preservation with varying sparsity levels, while ControlNet-based methods show inconsistencies in segmentation accuracy (highlighted in yellow boxes). \textit{ Captions: A stone building  surrounded by a stone wall and a grassy lawn.}
        }
    \label{ade20k}
\end{figure}
\begin{table}
    \centering
    \resizebox{\linewidth}{!}{
    \begin{tabular}{c|c|c|c|c} 
        \toprule[1.0px]
        \textbf{Method}  &\textbf{Base Model}& \textbf{FID $\downarrow$}& \textbf{CLIP\_score $\uparrow$}&\textbf{mIoU $\uparrow$}\\ 
        \midrule
        VQGAN \cite{esser2021taming} & \ding{55}& 26.28& 0.17& N/A \\ 
        LDM \cite{rombach2022high} & \ding{55}& 25.35& 0.18& N/A \\
        PIPT \cite{wang2022pretraining} & \ding{55}& 19.74& 0.20& N/A \\
        \midrule
        ControlNet &SD1.5& 21.33& 0.2531& 0.2764 \\
        ControlNet++ &SD1.5& 19.89& 0.2640& 0.3435 \\ 
        ControlNet-Large &SD1.5& 16.78& 0.2796&  0.3668 \\ 
        \midrule
        \rowcolor{gray!10}
        \textbf{FlexControl}$_{\gamma=0.3}$&SD1.5& 17.21& 0.2713& 0.3572 \\
        \rowcolor{gray!10}
        \textbf{FlexControl}$_{\gamma=0.5}$&SD1.5& \textcolor{blue}{14.80}& \textcolor{red}{0.2842}& \textcolor{blue}{0.3751} \\
        \rowcolor{gray!10}
        \textbf{FlexControl}$_{\gamma=0.7}$&SD1.5& \textcolor{red}{14.71}& \textcolor{blue}{0.2840}& \textcolor{red}{0.3775} \\ 
        \bottomrule[1.0px]
    \end{tabular}
    }
    \caption{
    \textbf{Quantitative comparison with existing methods on SD1.5.} 
    We report FID (↓), CLIP\_score (↑), and mIoU (↑) across different models. The best values and second-best are highlighted in \textcolor{red}{red} and \textcolor{blue}{blue}. FlexControl outperform original ControlNet with less computation, while increasing the blocks budgets observed performance increasing. Noticeable, ControlNet-Large activate all blocks yet not out-perform our methods, highlight effective of our dynamic strategy.
    }
    \label{tab:table4}
\end{table}

\subsection{Ablation study}

In this section, we analyze how the proportion of activated control blocks impacts FlexControl. To better understand model complexity, we present the number of parameters, FLOPs, and diffusion speed in \cref{tab:table3}. The diffusion iterations per second (\textit{i.e.}, it/s) for SD1.5-based models and seconds per iteration (\textit{i.e.}, s/it) for SD3.0-based models are measured on a single Nvidia RTX 2080 Ti GPU. We randomly select batch samples and compute the average single-step iteration time for each sample. 

\paragraph{Results on UNet-based model.}
Recall the cost loss defined in \cref{eq:losscost}, we train FlexControl with three different sparsity levels by adjusting the value of $\gamma$.  For the SD1.5-based backbone, experiments are conducted on the ADE20K dataset. At $\gamma=0.3$ ($~$30\% sparsity), FlexControl surpasses ControlNet and ControlNet++ in controllability and generation quality but falls short of ControlNet-Large. Increasing $\gamma$ to 0.5 activates more control blocks, leading to performance that surpasses ControlNet-Large. Further increasing $\gamma$ to 0.7 yields no significant performance gains (suggesting the dataset is already saturated by the model capacity).  For visual comparisons in \cref{ade20k}, FlexControl with $\gamma=0.5$ and $\gamma=0.7$ demonstrate superior structure preservation and mask information reconstruction. Meanwhile, the more lightweight configuration with $\gamma=0.3$ achieves a generation quality comparable to ControlNet++ and ControlNet-Large.

\begin{figure}[!t]
    \centering
        \includegraphics[width=1.0\columnwidth]{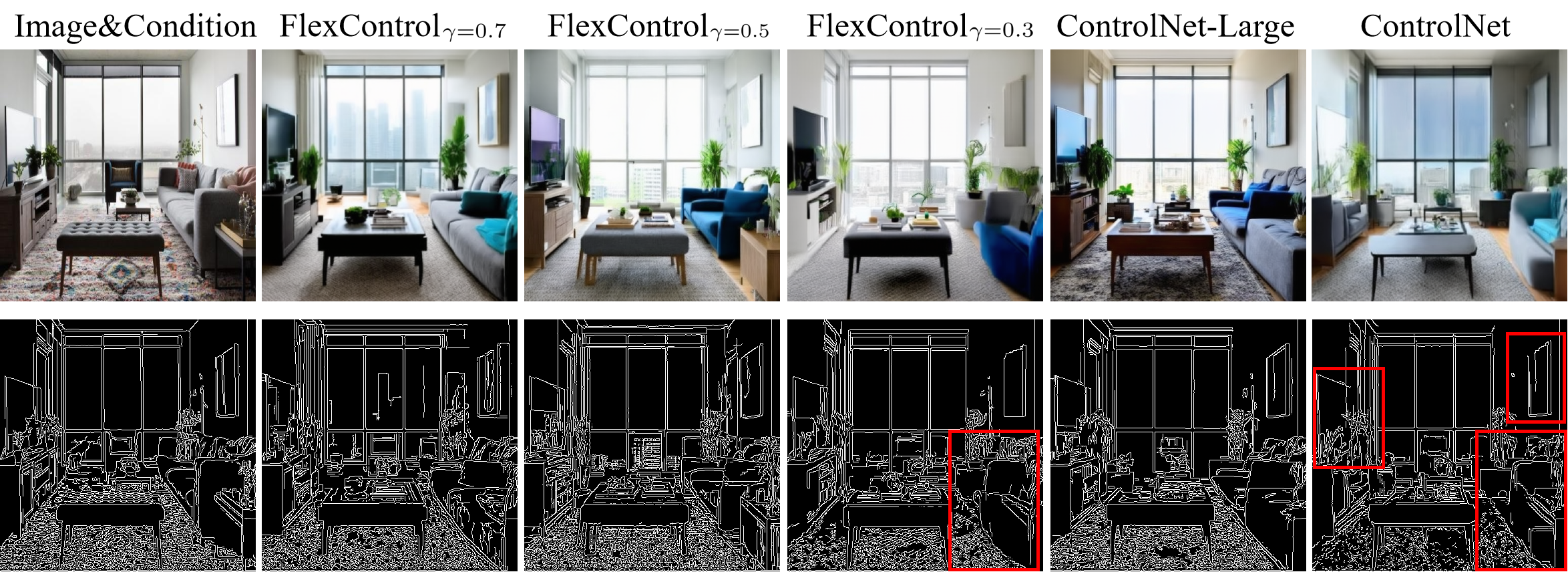}
        \caption{
        \textbf{Comparison of FlexControl and existing methods on SD3.0 for edge preservation.} 
        FlexControl maintains better spatial consistency and object integrity across different sparsity levels, while ControlNet-based methods introduce distortions and inconsistencies (highlighted in red boxes). \textit{Captions: A room with large windows, a gray sofa, a table, and a TV stand.}
        }

    \label{canny}
\end{figure}
\begin{table}
    \centering
    \resizebox{\linewidth}{!}{
    \begin{tabular}{c|c|c|c|c} 
        \toprule[1.0px]
        \textbf{Method}  &\textbf{Base Model}& \textbf{FID} $\downarrow$& \textbf{CLIP\_score} $\uparrow$&\textbf{SSIM} $\uparrow$\\ 
        \midrule
        ControlNet &SD3.0& 27.21& 0.2512& 0.3749\\
        ControlNet-Large &SD3.0& \textcolor{blue}{21.64}& \textcolor{blue}{0.2690}& \textcolor{red}{0.4828}\\
        \midrule
        \rowcolor{gray!10}
        \textbf{FlexControl}$_{\gamma=0.3}$ &SD3.0& 24.39& 0.2581& 0.4286\\
        \rowcolor{gray!10}
        \textbf{FlexControl}$_{\gamma=0.5}$ &SD3.0& 22.47& \textcolor{red}{0.2714}& 0.4598\\
        \rowcolor{gray!10}
        \textbf{FlexControl}$_{\gamma=0.7}$ &SD3.0& \textcolor{red}{20.54}& \textcolor{red}{0.2714}& \textcolor{blue}{0.4775}\\ 
        \bottomrule[1.0px]
    \end{tabular}
    }
    \caption{
        \textbf{Quantitative comparison with existing methods on SD3.0.} 
        We report FID (↓), CLIP\_score (↑), and SSIM (↑). 
        The best and second-best values are highlighted in \textcolor{red}{red} and \textcolor{blue}{blue}. FlexControl outperform original ControlNet with less computation, while increasing the blocks budgets observed performance increasing even more significant improvement than we observed in SD1.5.
        }
    \label{tab:table5}
\end{table}

\paragraph{Results on DiT-based model.}
For the SD3.0-based backbone, experiments were conducted on the LLAVA-558K dataset. As detailed in \cref{tab:table5}, FlexControl$_{\gamma=0.3}$ and FlexControl$_{\gamma=0.5}$ outperform ControlNet with fewer FLOPs. Notably, while ControlNet has half as many blocks as the backbone, each control block's output is shared by two adjacent backbone blocks, providing more conditional controls than FlexControl at all sparsity levels. FlexControl$_{\gamma=0.7}$ achieves superior image quality and comparable controllability to ControlNet-Large while being more efficient. Visualization results in \cref{canny} further demonstrate the advantage of FlexControl in edge reproduction and image fidelity over ControlNet.

%% file: sec/5_conclusion.tex
\section{Conclusion}

We presented FlexControl, a dynamic framework that reimagines controlled diffusion by replacing heuristic block selection with a trainable, computation-aware gating mechanism. By enabling adaptive activation of control blocks during denoising, FlexControl eliminates manual architectural tuning, reduces computational overhead, and maintains or improves output fidelity across diverse tasks and architectures (UNet, DiT). Our experiments demonstrate that flexibility and efficiency need not be mutually exclusive in controllable generation—intelligent, data-driven activation strategies can outperform rigid, handcrafted designs. This work paves the way for future research into lightweight, generalizable control mechanisms for increasingly complex generative pipelines. We also conduct a further investigation on the dynamic activation route, which has been listed in \cref{app:exploration}.

%% file: sec/6_appendix.tex
\NewDocumentCommand{\csq}{om}{%
  % colored square
  % #1 = color model (optional)
  % #2 = color spec
  \IfNoValueTF{#1}{%
    \textcolor{#2}{\rule{1.2ex}{1.2ex}}%
  }{%
    \textcolor[#1]{#2}{\rule{1.2ex}{1.2ex}}%
  }%
}

\clearpage

\renewcommand{\thefigure}{A\arabic{figure}}
\setcounter{figure}{0}
\renewcommand{\thetable}{A\arabic{table}}
\setcounter{table}{0}
\renewcommand{\thesection}{A\arabic{section}}
\setcounter{section}{0}
\maketitlesupplementary

%\onecolumn
The supplementary material presents the following sections to strengthen the main manuscript:

\begin{itemize}
\item Pseudocode of our algorithm.
\item Implementation details.
\item Dynamic route exploration.
\item More visualization
\end{itemize}

\section{Pseudocode of Our Algorithm}
\label{app:pseudo}

\begin{algorithm}[!ht]
    \caption{Inference procedure}
    \label{alg:inference}
    \renewcommand{\algorithmicrequire}{\textbf{Input:}}
    \begin{algorithmic}[1]
        \REQUIRE conditional image $\mathbf{c}_{s}$, text prompt $\mathbf{c}_{t}$, timestep $T$. \\ Fully-trained FlexControl, pre-trained SD model.%%input
        %\ENSURE out    %%output
        \FOR{each $i \in [T,1]$}
            \FOR{each $l \in Blocks$}
               \STATE \textcolor{magenta}{/$^{*}$ The value of $\mathcal{M}_{l}$ is adjusted with input $\mathbf{h}_{l-1}$ $^{*}$/}
               \STATE Compute $\mathcal{M}_{l}$ though router unit
                \IF{$\mathcal{M}_{l}=1$}
                   \STATE \textcolor{magenta}{/$^{*}$ Extract latent features from conditions $^{*}$/}
                   \STATE Compute $\mathbf{h}_{l}$ though \cref{eq:forward_backward_1} 
                   \STATE \textcolor{magenta}{/$^{*}$ Feature transformation by zero modules $^{*}$/}
                   \STATE Transform $\mathbf{h}_{l}$ to $\mathbf{y}_{c}^{l}$ though \cref{eq:forward_backward_2}
                   \STATE \textcolor{magenta}{/$^{*}$ Inject modal information into feature space $^{*}$/}
                   \STATE Inject $\mathbf{y}_{c}^{l}$ to SD model though \cref{eq:inject_conditional_controls}
                \ELSE
                   \STATE \textcolor{magenta}{/$^{*}$ Align the dimension of feature mapping $^{*}$/}
                   \STATE Bypass $\mathcal{F}_{l}$ and $\mathcal{Z}_{l}$ though $\mathrm{skip}_{l}\left(\cdot\right)$  %\Comment{This is a comment}
                \ENDIF
            \ENDFOR
            \STATE \textcolor{magenta}{/$^{*}$ DDIM sampler for SD1.5-based models $^{*}$/}
            \STATE \textcolor{magenta}{/$^{*}$ RFlow sampler for SD3.0-based models $^{*}$/}
            \STATE Predict denoised image with $T$-step sampling
        \ENDFOR
        \STATE \textbf{return} $\mathbf{x}_{0}$
    \end{algorithmic}
\end{algorithm}

\begin{algorithm}[!ht]
    \caption{Training procedure}
    \label{alg:training}
    \renewcommand{\algorithmicrequire}{\textbf{Input:}}
    \renewcommand{\algorithmicensure}{\textbf{Output:}}
    \begin{algorithmic}[1]
        \REQUIRE Dataset $\mathcal{D}\left(x,\mathbf{c}_{s},\mathbf{c}_{t}\right)$, hyperparameters $\left(\tau,\gamma,\lambda_{\mathbf{C}}\right)$. \\ Initialized FlexControl model, pre-trained SD model.%%input
        %\ENSURE out    %%output
        \STATE Turn off the router units for warm-up training
        \STATE Turn on the router units for end-to-end training
        \WHILE{not converged}
            \STATE Sample timestep $t\sim Uniform\left(\mathbf{0},\mathbf{1}\right)$
            \STATE Sample nosie $\epsilon\sim\mathcal{N}\left(\mathbf{0},\boldsymbol{\mathit{I}}\right)$
            \STATE \textcolor{magenta}{/$^{*}$ \cref{eq:ddpm_forward} is used for SD1.5-based models $^{*}$/}
            \STATE \textcolor{magenta}{/$^{*}$ \cref{eq:fm_forward} is used for SD3.0-based models $^{*}$/}
            \STATE Transfer image $\mathbf{x}_{0}$ to noisy image $\mathbf{x}_{t}$
            \STATE \textcolor{magenta}{/$^{*}$ Based on \cref{eq:redefined_forward_backward_1,eq:redefined_forward_backward_2,eq:inject_conditional_controls} $^{*}$/}
            \STATE $\mathbf{y}_{pred},\mathcal{M}=\hat{\epsilon}_{\theta}\left(\mathbf{x}_{t},\mathbf{c}_{t},\mathbf{c}_{s},t\right)$
            \STATE \textcolor{magenta}{/$^{*}$ $\mathcal{L}_{\mathbf{SD}}$ is used to optimize generation effect $^{*}$/}
            \STATE Compute MSE loss $\mathcal{L}_{\mathbf{SD}}$ though \cref{eq:losssd}
            \STATE \textcolor{magenta}{/$^{*}$ $\mathcal{L}_{\mathbf{C}}$ is used to control sparsity $^{*}$/}
            \STATE Compute cost loss $\mathcal{L}_{\mathbf{C}}$ though \cref{eq:losscost}
            \STATE \textcolor{magenta}{/$^{*}$ $\mathcal{L}_{\theta}$ is used as final optimization goal $^{*}$/}
            \STATE Compute final loss $\mathcal{L}_{\theta}$ though \cref{eq:losstheta}
            \STATE \textcolor{magenta}{/$^{*}$ Freeze the weight parameters of the backbone $^{*}$/}
            \STATE $\theta=\theta-lr\nabla_{\theta}\mathcal{L}_{\theta}\left(\mathbf{x}_{t},\mathbf{y}_{pred},\mathcal{M}\right)$
        \ENDWHILE
        \STATE \textbf{return} fully-trained FlexControl
    \end{algorithmic}
\end{algorithm}

\section{Implementation Details}
We implement FlexControl based on SD1.5 \cite{SD1.5} and SD3.0 \cite{SD3}. The experiments are carried out under various conditions, mainly including depth map, canny edge and segmentation mask. The following is a description of the experimental details.

\paragraph{Training dateset.}
The experiment involves three types of conditional maps:
\begin{itemize}
\item[$\bullet$] \textit{Depth map.} In this application, we use MultiGen-20M proposed by \cite{zhao2024uni} as training data, which is a subset of LAION-Aesthetics \cite{schuhmann2022laion} and contains over 2 million depth-image-caption pairs, and 5K test samples.
\item[$\bullet$] \textit{Canny edge.} For the condition of the canny edge, we use the LLAVA-558K \cite{liu2024visual} dataset to verify the model, which contains 558K image-caption pairs. A canny edge detector \cite{canny1986computational} is used to convert RGB images to edge images, and the low and high threshold of hysteresis procedure in this process are set to 100 and 200, respectively. 
\item[$\bullet$] \textit{Segmentation mask.} For the segmentation mask, we use the ADE20K \cite{zhou2017scene} dataset for model training. This dataset contains a total of 27K segmentation image pairs, 25K for training and 2K for testing. InternVL2-2B \cite{chen2024internvl} is used to generate captions for RGB images with instruction ``Please use a brief sentence with as few words as possible to summarize the picture”.
\end{itemize}

\paragraph{Training settings.}
During the training procedure, we uniformly use the AdamW optimizer with a learning rate of 1$\times$10$^{-5}$. For SD1.5-based models, half-precision floating-point (Float16) is used for mixed precision training, original images and conditional images are resized to 512$\times$512, and batchsize and gradient accumulation steps are set to 4 and 32, respectively. When turning to SD3.0, we further use DeepSpeed \cite{rajbhandari2020zero} Zero-2 to accelerate the training process, the resolution of 1024$\times$1024 is used, and the batchsize and gradient accumulation steps are set to 4 and 8. We set the maximum training iterations to 50k and 25k for the models based on SD1.5 and SD3.0, respectively. For the threshold parameter $\tau$ required by the Gumbel-Sigmoid activation function in the router unit, we set it to 0.5, and the hyperparameter $\lambda_{\mathbf{C}}$ in the loss function $\mathcal{L}_{\theta}$ is set to 0.5, the value of $\gamma$ depends on the target sparsity. When training UNet-based ControlNet-large and FlexControl, we remove the residual connection between the encoder blocks and the decoder blocks of the control network. For the problem of the weight dimension cannot be aligned when initializing the decoder blocks of the control network using SD1.5's pre-trained weights caused by this operation, we solve it by reinitializing these weights. The models based on SD1.5 and SD3.0 are trained with 2 and 8 Nvidia-A100 (40G) GPUs, respectively.

Our FlexControl follows the core design philosophy of \cite{zhang2023adding}, the trainable blocks are initialized with the pre-trained weight parameters of the SD model, and zero modules are added at the same time, which leads to the conditional mappings generated at the early training stage do not have the ability to control generation effectively. Therefore, we first fix mask $\mathcal{M}$ to 1 for warm-up training in the early training stage, \textit{e.g.}, 10K steps for SD1.5-based FlexControl and 5K steps for SD3-based FlexControl in our implementation, and then turn on the router unit to train together with the copy blocks. This helps the whole training procedure move in the right direction.

\paragraph{Benchmark and metrics.}
For quantitative comparison, we present the Frechet Inception Distance (FID) \cite{heusel2017gans} and CLIP\_score \cite{radford2021learning} to assess the quality of the generated images. In addition, we calculate RMSE, SSIM and mIoU on depth map, canny edge and segmentation mask respectively, to evaluate the controllability of image generation. Finally, we emphatically compare the computational complexity. The results of depth map are tested on MultiGen-20M test set and the results of canny edge and segmentation mask are tested on COCO validation set, which contains 5,000 samples and each sample contains five text descriptions, we randomly choose one text of each sample as input during testing. For sampling, we employ DDIM \cite{song2020denoising} and RFlow \cite{SD3} sampler, implementing 20 denoising steps to generate images without incorporating any negative prompts. We generate five groups of images, and the average results are reported.
%RFlow with dynamic rectification strategy

%\section{More Experiments about Dynamic Route}
\section{Dynamic Route Exploration}
\label{app:exploration}

In order to improve the parameter utilization of ControlNet in the application, we explore how the router unit activates the control block to generate conditional controls.

It can be seen from \cref{fig1}(c), both SD1.5 based on UNet and SD3.0 based on DiT, the activation of control blocks presents a sparse distribution in the early denoising stage and a dense distribution in the late stage. This means that the late denoising stage plays a more important role in controllable image generation. Since the early sampling is mainly responsible for generating the global structure and low-frequency information of the image (\textit{e.g.}, the approximate shape of the object, the distribution of the components), while the late sampling is mainly responsible for generating high-frequency information and correcting complex details (\textit{e.g.}, edge, texture). For this, more conditional control signals are necessary. Moreover, the generation deviation in the early stage is relatively small and can be rectified by subsequent sampling. If the sampling error in the later stage is large, the conditional consistency will be destroyed, resulting in loss of control effect. 

Based on the above findings, we can conclude that using unified control scheme in any case is an inefficient control mode, which leads to most of the conditional controls added in the early stage not playing the ideal role, and there will be insufficient conditional controls added in the late stage. Therefore, our dynamic control method can further release the performance of the controllable generation model by solving this problem. Next, we do a more detailed analysis of the different settings.

\begin{figure*}[!t]
    \centering
    \includegraphics[width=1.0\textwidth]{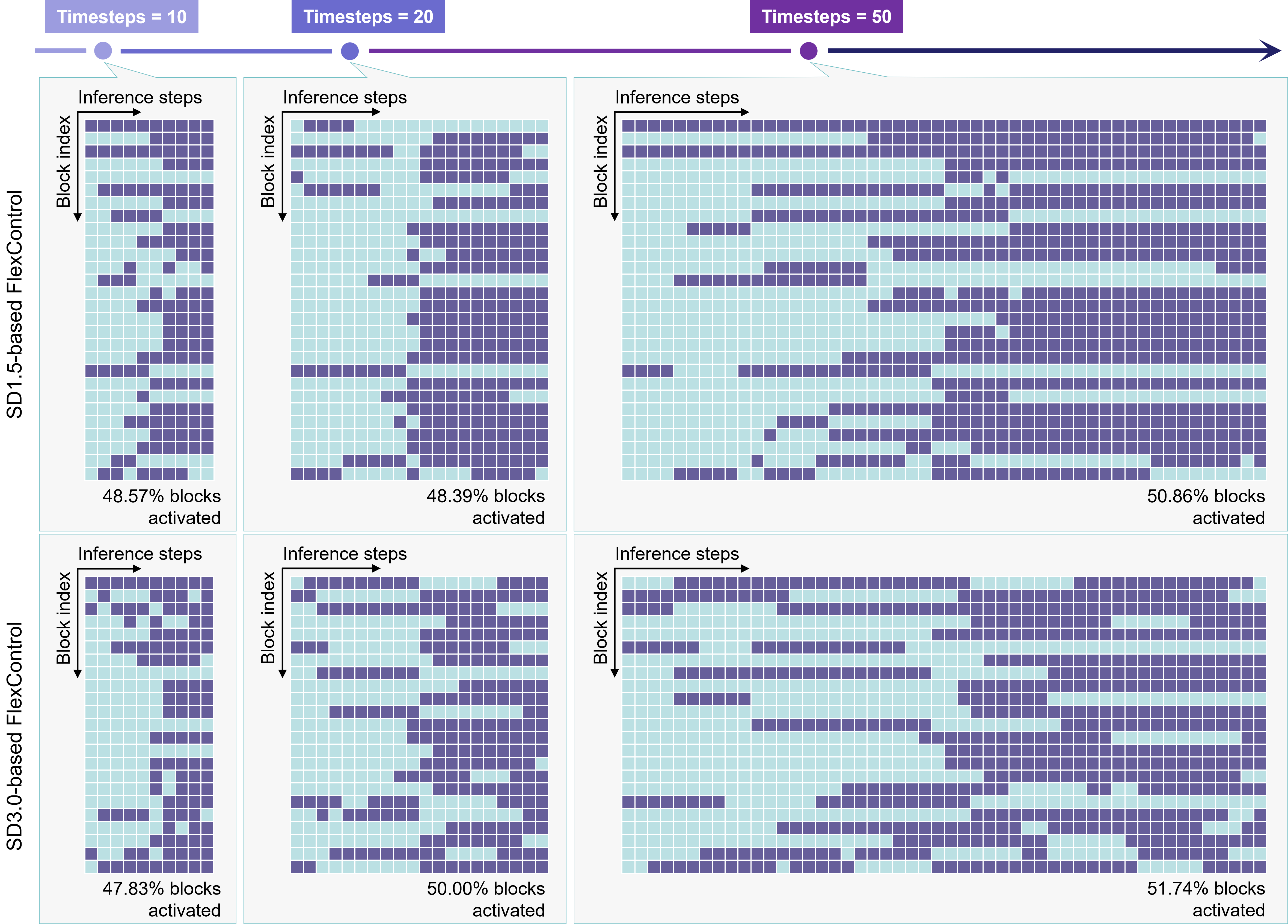}
    \caption{\textbf{The distribution of activated control blocks under different timesteps.} The hyperparameter $\gamma$ is set to 0.5 to approximate 50\% sparsity, and timestep is set to 10, 20 and 50 repectively. The first line shows the results of the model based on SD1.5, and the second line shows the results of the model based on SD3.0. $\large\csq[HTML]{665E9A}$ and $\large\csq[HTML]{BBE0E3}$ denotes activated and inactivated blocks, respectively.}
    \label{inf_steps}
\end{figure*}

First, we test the activation time and position of the control block under different number of timestep settings with $\gamma$ set to 0.5 (\textit{i.e.}, approximately 50\% sparsity). We set the timesteps to 10, 20 and 50 respectively. As shown in \cref{inf_steps}, it can be found that the pattern under different settings is basically the same as above. In the early stage, only a few blocks are activated, mainly concentrated in the head and tail. As the number of sampling steps increases, more blocks are activated. Until the middle stage of sampling, most of the blocks are activated.

\begin{figure*}[!t]
    \centering
    \includegraphics[width=1.0\textwidth]{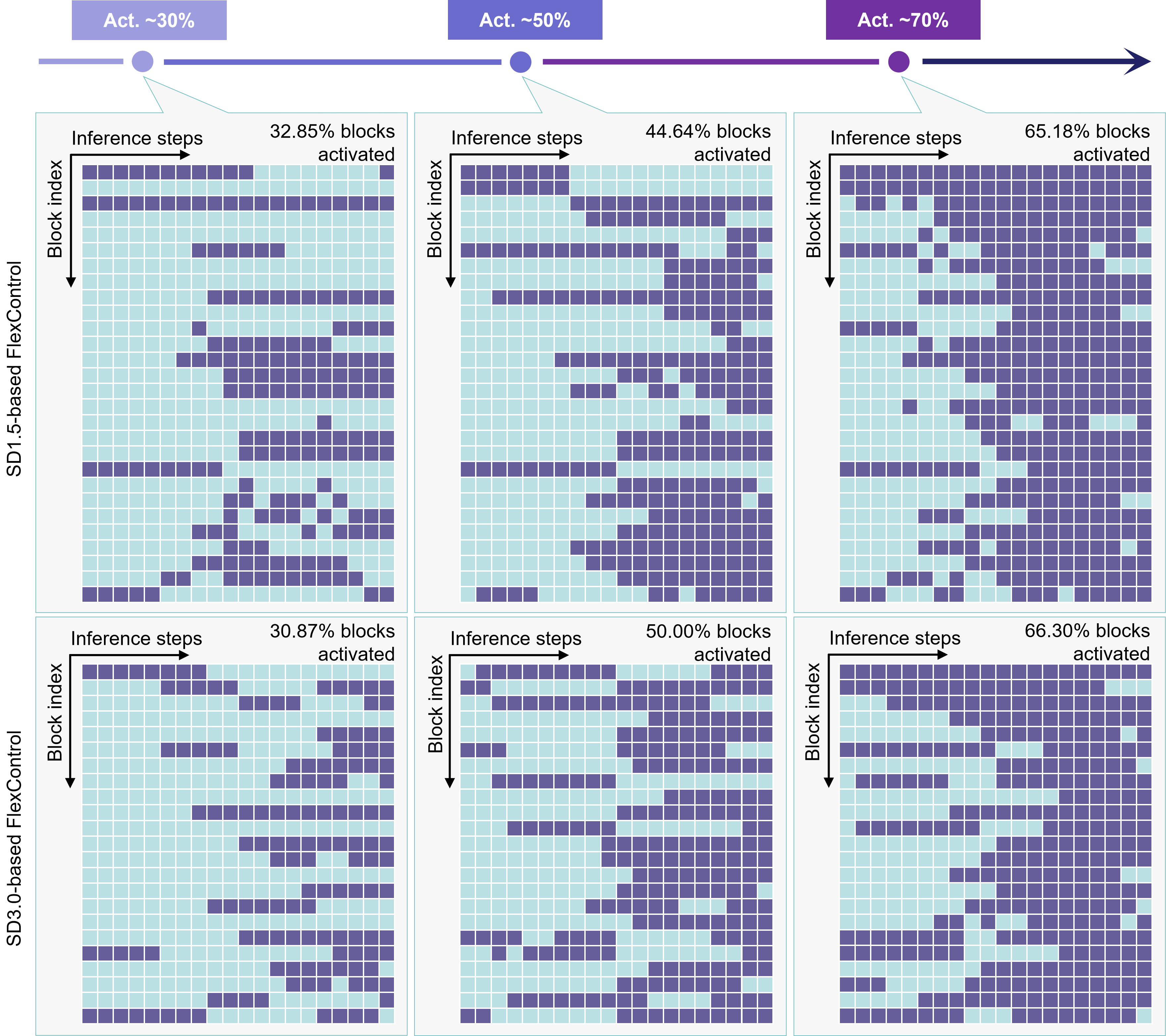}
    \caption{\textbf{The distribution of activated control blocks under different sparsity.} The hyperparameter $\gamma$ is set to 0.3, 0.5 and 0.7 to approximate 30\%, 50\%, and 70\% sparsity, and the timestep is set to 20. The first line shows the results of the model based on SD1.5, and the second line shows the results of the model based on SD3.0.}
    \label{act_inf_steps}
\end{figure*}

Next, we test the activation of control blocks under different sparsity. We approximate 30\%, 50\%, and 70\% sparsity by setting $\gamma$ to 0.3, 0.5, and 0.7. It can be seen from \cref{act_inf_steps}, when 30\% sparsity is used, fewer blocks are activated at the late stage, and even some control blocks are not activated at all. When the sparsity increased to 70\%, more middle blocks are activated in the early sampling period, and almost all blocks are activated in the late sampling period.

\begin{figure*}[!t]
    \centering
    \includegraphics[width=1.0\textwidth]{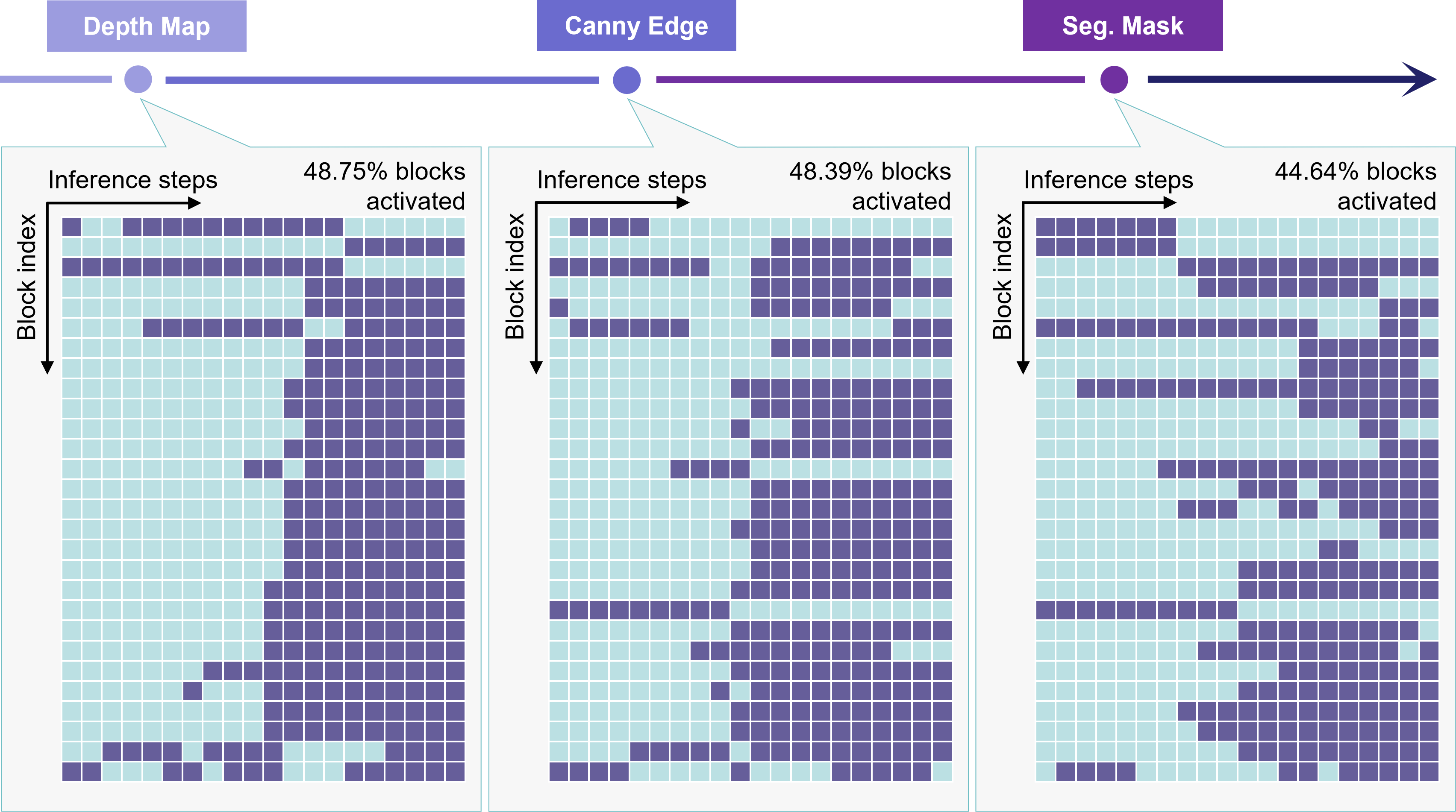}
    \caption{\textbf{The distribution of activated control blocks under various conditional controls.} The hyperparameter $\gamma$ is set to 0.5 to approximate 50\% sparsity, and the timestep is set to 20.}
    \label{con_inf_steps}
\end{figure*}

In addition, we discuss the activation of various spatial conditions at each timestep. As shown in \cref{con_inf_steps}, similar trend is found across different types of conditional maps, which proves the generalization of the above findings. In addition, there are some differences in activation details, which means that the router unit makes independent judgments on different conditional samples and plans specific activation routes for them. Due to the differences in feature distribution and information in the samples, this fine-grained control is particularly important for striking a balance between performance and efficiency.

Relying on the above findings, when we apply ControlNet or similar architectures in practice, only activating the head and tail blocks in the early stage, or even activating ControlNet only in the late stage, can simply improve the inference efficiency, and no retraining is involved.

\section{More Visualization}

\begin{figure*}[!t]
    \centering
	\includegraphics[width=1.0\textwidth]{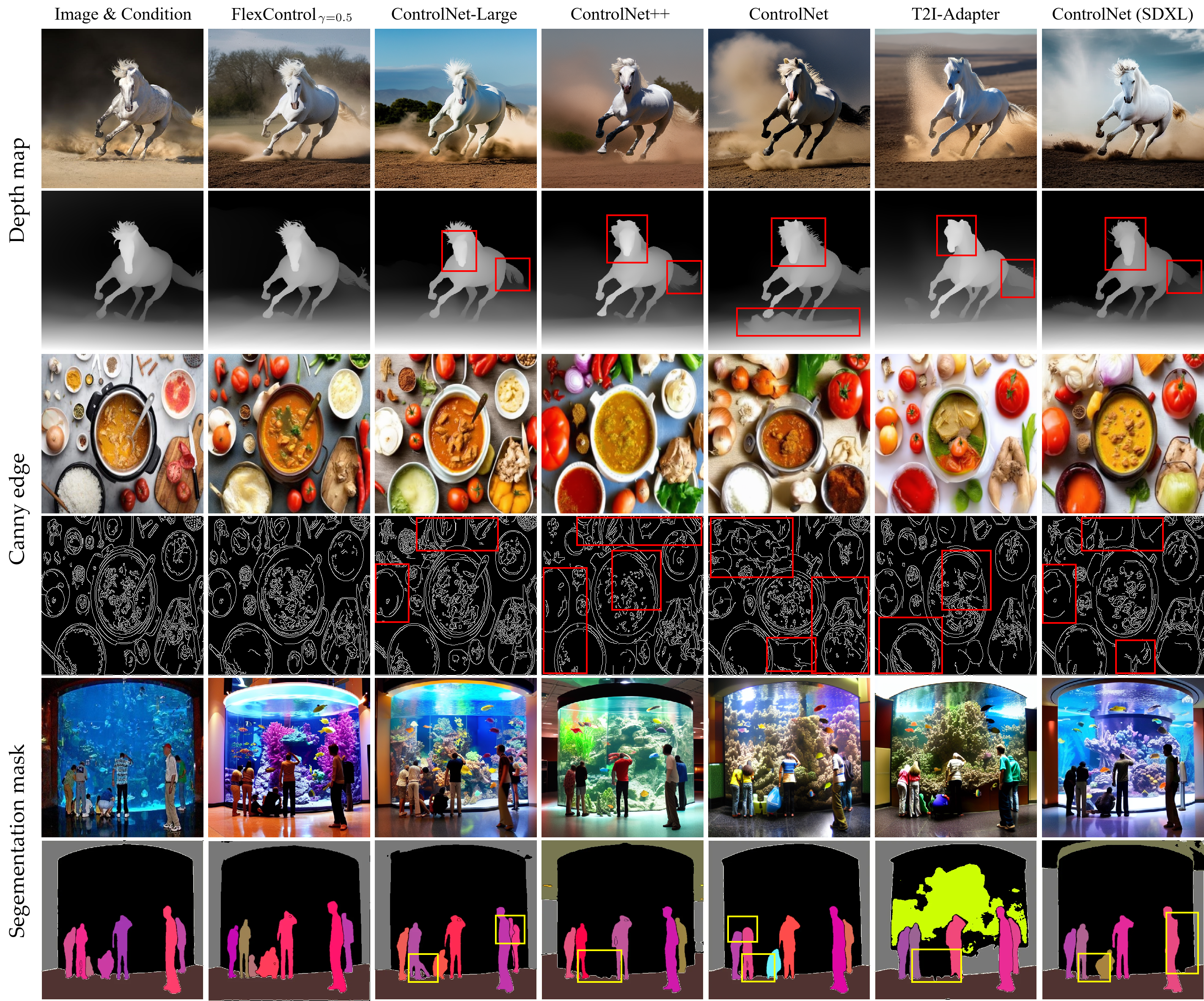}
	\caption{\textbf{Visualization comparison} with state-of-the-art controllable generation methods on various conditions. Except for the last two columns, the other models use SD1.5 as the backbone. \textit{Captions: A white stallion horse galloping furiously kicking up the dust behind it. Ingredients of curry, including onions, garlic, chili, and tomatoes. A group of people are observing an aquarium filled with colorful fish.}}
    \label{app_all_conditions}
\end{figure*}

\begin{figure*}[!t]
    \centering
	\includegraphics[width=1.0\textwidth]{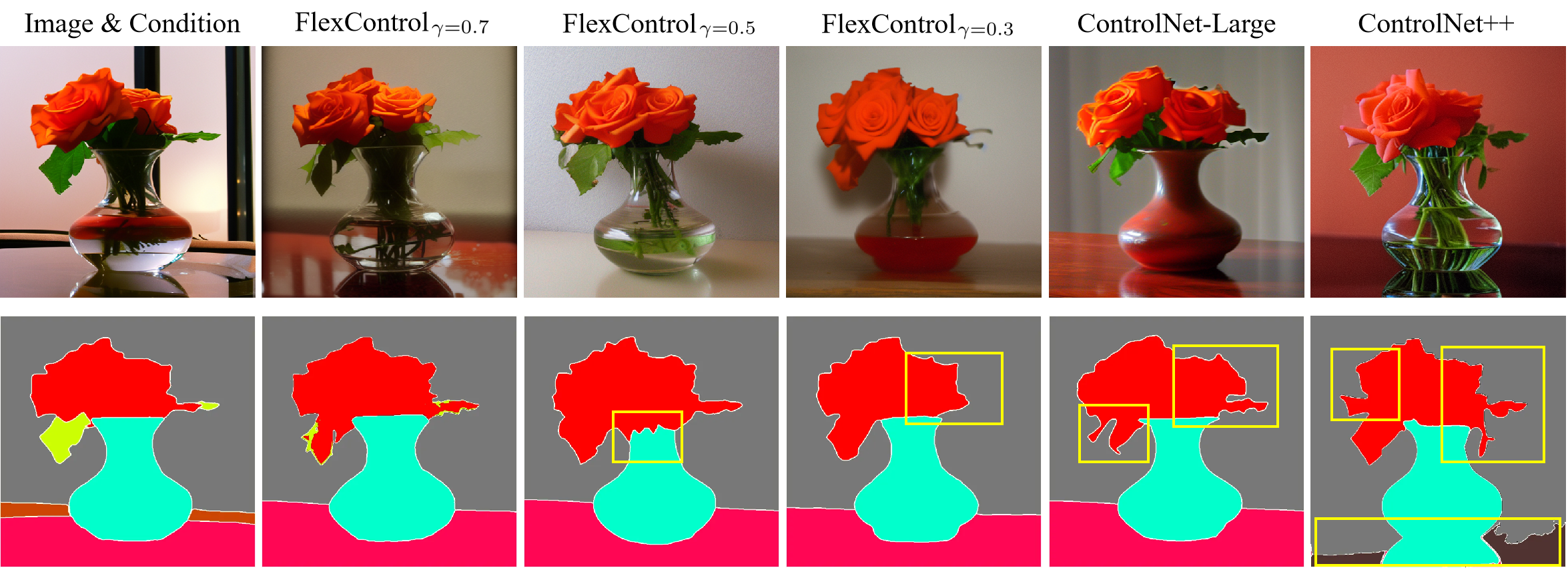}
	\caption{\textbf{Visualization comparison} of FlexControl and existing methods on SD1.5 for semantic consistency. \textit{Captions: A reddish rose in a vase filled with water on the table.}}
    \label{app_seg_mask}
\end{figure*}

\begin{figure*}[!t]
    \centering
	\includegraphics[width=1.0\textwidth]{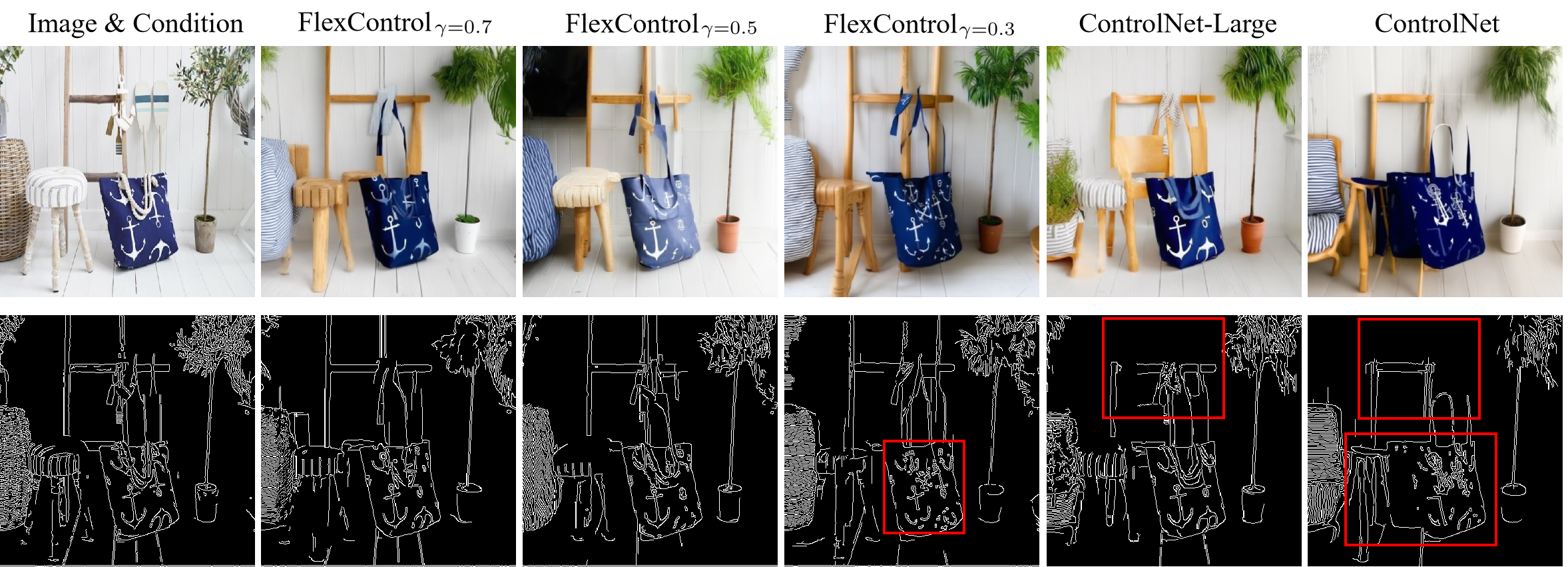}
	\caption{\textbf{Visualization comparison} of FlexControl and existing methods on SD3.0 for edge preservation. \textit{Captions: A wooden chair with a striped cushion, a navy blue tote bag with anchors, and a potted plant are arranged on a white floor against a white wooden wall.}}
    \label{app_canny}
\end{figure*}